\documentclass[letterpaper, 10 pt, conference]{ieeeconf}  

\IEEEoverridecommandlockouts                              

\usepackage{multicol}
\usepackage{multirow}
\usepackage[usenames,dvipsnames]{xcolor}
\usepackage[colorlinks=true, citecolor=blue, linkcolor=black, urlcolor=blue]{hyperref}

\usepackage{times} 

\usepackage{graphicx}
\usepackage{algorithm}
\usepackage{algpseudocode}
\usepackage{amsmath}
\usepackage{amssymb}  
\usepackage{pifont}
\usepackage{nicefrac}       
\usepackage[normalem]{ulem}
\usepackage{capt-of}
\usepackage{booktabs}
\usepackage[table]{xcolor}
\usepackage{pifont}

\newcommand{\cmark}{\textcolor{blue!65!black}{\ding{51}}}
\newcommand{\xmark}{\textcolor{red!65!black}{\ding{55}}}

\title{\LARGE \bf
Build Once, Monitor Continuously: Persistent Semantic Mapping via \\ Autonomous Exploration and Open-Vocabulary Object Updates

}

\author{Sai Haneesh Allu, Itay Kadosh, Tyler Summers, Yu Xiang
\thanks{Sai Haneesh Allu, Itay Kadosh and Yu Xiang are with the Department of Computer Science, and Tyler Summers is with the Department of Mechanical Engineering, University of Texas at Dallas, Richardson, TX 75080, USA \tt\small \{saihaneesh.allu, itay.kadosh, tyler.summers, yu.xiang\}@utdallas.edu} 
}

\usepackage{array}
\newcolumntype{N}{>{\centering\arraybackslash}m{3.5cm}}
\newcolumntype{M}{>{\centering\arraybackslash}m{2.5cm}}

\begin{document}


\maketitle
\thispagestyle{empty}
\pagestyle{empty}


\begin{abstract}
Persistent semantic monitoring of indoor spaces such as warehouses, hospitals, and offices requires a robot to repeatedly monitor an environment and track how objects change over time. Running full simultaneous localization and mapping (SLAM) with dense semantic reconstruction from scratch on every visit is redundant when the environment geometry stays the same and only the objects move. We present a modular two-stage system that separates geometric mapping from semantic updating. In the first stage, a frontier-based exploration method with a dynamic search window builds a 2D occupancy grid. In the second stage, the robot relocalizes in this map and builds a semantic object graph using an open-vocabulary object detector and a promptable segmentation model. Only the lightweight semantic stage is repeated on later visits, so the system scales well to frequent revisits. The object graph uses a category and distance based association rule to update objects, which lets the map reflect both intra-session changes (object changes within a single traversal) and inter-session changes (changes across revisits), such as objects being moved, removed, or added. We validate the system on a Fetch robot in two real indoor environments of about $8{,}500$\,m$^2$ and $117$\,m$^2$, and report precision, recall, and F1 scores across multiple update iterations.\footnote{Project page with code is available at: \url{https://irvlutd.github.io/SemanticMapping}}

\end{abstract}


\section{Introduction}

Autonomous exploration of unknown environments is an important problem in robotics, with applications in search and rescue~\cite{queralta2020collaborative}, surveillance~\cite{chun2016robot}, and service robotics~\cite{belanche2020service}. While significant progress has been made in building geometric maps using LiDAR-based occupancy grids~\cite{thrun2003learning,yamauchi1997frontier} and 3D representations with RGB-D SLAM~\cite{mur2015orb,newcombe2011kinectfusion}, these maps lack semantic understanding of the objects within them.

Adding semantics to geometric maps, often called \emph{semantic mapping}~\cite{kostavelis2015semantic}, is important for downstream tasks such as object-goal navigation~\cite{chaplot2020object,cai2024bridging}. Recent advances leverage open-vocabulary detectors~\cite{groundingdino} and foundation models to build rich scene representations~\cite{gu2024conceptgraphs,hughes2022hydra,maggio2024clio}. However, most semantic mapping systems build a map once and do not support updating it to reflect environment changes such as object relocation or removal.

In many practical deployments, such as warehouses, hospitals, and office buildings, a robot must monitor the same environment \emph{repeatedly}. Objects are moved, added, or removed on a daily basis, and the semantic map has to be updated to match. Running full SLAM with dense semantic reconstruction from scratch on every visit repeats a lot of work, since the building geometry stays largely the same and only the objects change. This motivates a \emph{two-stage} design: build a reliable geometric map once, then relocalize and update only the semantic layer on each later visit. We refer to this idea as \textbf{Build Once, Monitor Continuously}.

\begin{figure}[t]
\vspace{3mm}
    \centering
\includegraphics[width=\linewidth]{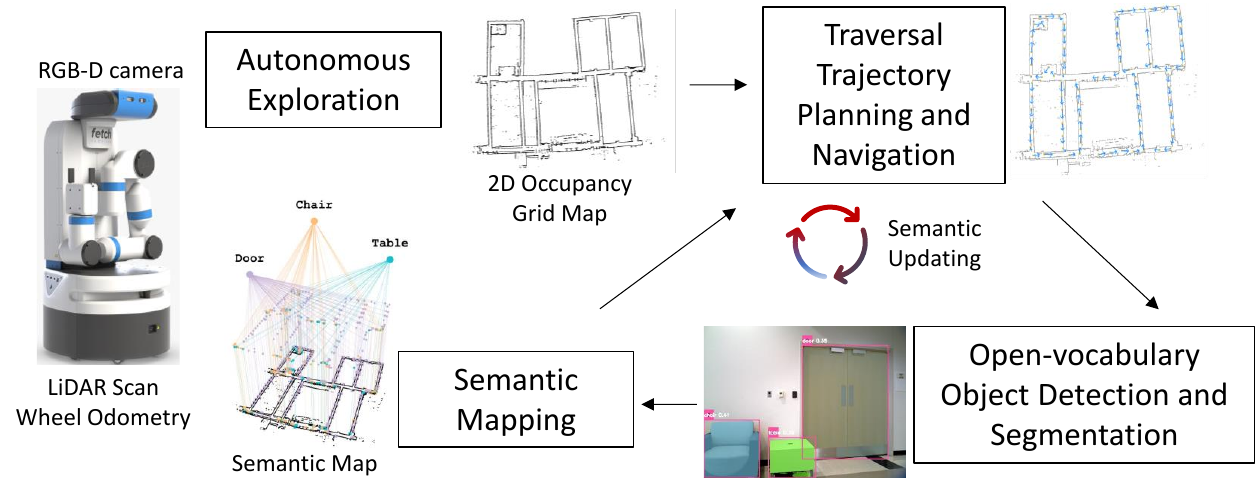}
    \caption{Our system enables a mobile robot to explore a large-scale unknown environment, build a semantic map and update the map for environment changes.}
    \label{fig:intro}
    \vspace{-1mm}
\end{figure}

Our system implements this paradigm on a physical robot. In the first stage, a modified frontier-based exploration method builds a 2D occupancy grid. In the second stage, the robot relocalizes within the built map and traverses the environment along a planned trajectory, using an open-vocabulary object detector and a promptable segmentation model to build and update a \emph{semantic object graph}. The graph captures both \emph{intra-session changes}, where objects change during a single traversal after the robot has already passed that section, and \emph{inter-session changes} that build up across separate revisits. We present this work as a systems contribution rather than a single new algorithm. Our focus is the persistent monitoring setting, where efficient and reliable deployment over many revisits matters more than dense single-pass reconstruction. We validated the system in a large ($8{,}500$\,m$^2$) and a medium ($117$\,m$^2$) indoor environment using a Fetch robot with multiple update iterations. We plan to release the code and recorded data on publication.

 Our main contributions are as follows.
\begin{itemize}
    \item \textbf{An integrated and fully autonomous system}: We bring together autonomous exploration, relocalization, open-vocabulary semantic mapping, and map updating in a single pipeline that runs on a physical robot without human teleoperation in either stage. Within this pipeline we use a lightweight object graph and a category and distance based association rule to add, move, and remove objects within one visit and across revisits.
    \item \textbf{A decoupled two-stage design with an explicit trade-off}: The geometric map is built once, and only the lightweight semantic stage is repeated on later visits. We explain why this is a reasonable choice for persistent monitoring, where the system trades dense reconstruction for fast and repeatable object-level updates that scale with the number of revisits.
    \item \textbf{Real-world validation and analysis}: We deploy and evaluate the system on a Fetch robot in a large ($8{,}500$\,m$^2$) and a medium ($117$\,m$^2$) indoor environment, report precision, recall, and F1 across multiple update iterations, and compare object-level mapping throughput against a state-of-the-art baseline.
\end{itemize}

\section{Related Work}

\subsection{Autonomous Exploration}

Frontier-based exploration, introduced by Yamauchi~\cite{yamauchi1997frontier}, is still the most common approach for autonomous robot exploration. Common extensions use information gain to select frontiers~\cite{keidar2014efficient,faigl2015benchmarking,dai2020fast}, or use hierarchical local-global planners such as FUEL~\cite{zhou2021fuel}, TARE~\cite{cao2021tare}, and the unified volumetric planner of Schmid et al.~\cite{schmid2021unified} to cover large environments quickly. A separate line of work adds semantics to exploration. Some methods use object detections or semantic objectives to choose frontiers~\cite{asgharivaskasi2023semantic,zhang2024active} or use vision-language models for zero-shot navigation~\cite{yokoyama2024vlfm}, while more recent work brings in foundation models, learning to imagine unseen regions~\cite{shah2025foresightnav} or keeping open-set semantic frontiers for online exploration~\cite{alama2025rayfronts}. We note a distinction between using semantics to \emph{guide} exploration and building a semantic map \emph{while} exploring; our system targets the latter. Our exploration module is a simple frontier method with a dynamic search window. It only needs to give good geometric coverage, and it can be swapped for any of the planners above without changing the rest of the system.

\subsection{Semantic Mapping and Scene Graphs}

A large body of work builds semantic maps from posed RGB-D data, fusing semantic labels into 3D. Early methods combine dense reconstruction with learned segmentation~\cite{mccormac2017semanticfusion,xiang2017rnn,narita2019panopticfusion}. More recent work uses hierarchical scene graphs~\cite{hughes2022hydra,hughes2024foundations} and open-vocabulary detectors such as GroundingDINO~\cite{groundingdino} for richer, category-agnostic maps. ConceptGraphs~\cite{gu2024conceptgraphs} and its hierarchical extension~\cite{Werby-RSS-24} build object-centric 3D scene graphs from pre-collected scans, Clio~\cite{maggio2024clio} builds real-time, task-driven open-set scene graphs onboard a robot, and Kimera2~\cite{abate2024kimera2robustaccuratemetricsemantic} and Open-Fusion~\cite{yamazaki2024open} target dense semantic reconstruction. Neural mapping~\cite{kuang2024activeneuralmappingscale} uses NeRF or Gaussian representations for scalable active mapping, and SayNav~\cite{arajv2024Saynav} uses large language models for navigation planning. These methods build expressive maps, but they generally build the map once and do not update it as objects change. Neural field representations also have a computational cost that limits frequent, real-time revisits on mobile robots. We instead focus on the complementary problem of \emph{maintaining} a semantic map across repeated visits.

\begin{table}[t]
\vspace{2.5mm}
\centering
\renewcommand{\arraystretch}{2.3}

\rowcolors{2}{gray!12}{white}

\resizebox{\linewidth}{!}{
\begin{tabular}{lcccc}
\toprule
\rowcolor{gray!30}
\textbf{Method} &
\textbf{Autonomous} &
\textbf{Metric Mapping} &
\textbf{Semantic} &
\textbf{Semantic} \\

\rowcolor{gray!30}
&
\textbf{Exploration} &
\textbf{with Robot} &
\textbf{Mapping} &
\textbf{Updating} \\
\midrule

SemanticFusion~\cite{mccormac2017semanticfusion}
& \xmark & \xmark & \cmark & \xmark \\

DA-RNN~\cite{xiang2017rnn}
& \xmark & \xmark & \cmark & \xmark \\

Hydra~\cite{hughes2022hydra}
& \xmark & \xmark & \cmark & \xmark \\

ConceptGraphs~\cite{gu2024conceptgraphs}
& \xmark & \xmark & \cmark & \xmark \\

Dengler et al.~\cite{dengler2021online}
& \xmark & \xmark & \cmark & \cmark \\

Khronos~\cite{schmid2024khronosunifiedapproachspatiotemporal}
& \xmark & \cmark & \cmark & \cmark \\

\rowcolor{gray!18}
\textbf{Ours}
& \textbf{\cmark}
& \textbf{\cmark}
& \textbf{\cmark}
& \textbf{\cmark} \\

\bottomrule
\end{tabular}
}

\caption{Comparison of robotic exploration and mapping paradigms.}
\label{tab:method-compare}
\vspace{-5mm}
\end{table}

\subsection{Change Detection and Long-Term Mapping}
Detecting and representing environmental changes is essential for robots operating over extended periods. Dengler et al.~\cite{dengler2021online} construct and update a semantic map in real-time using instance segmentation on a mobile manipulator, though their system requires a pre-built map and manual control. Khronos~\cite{schmid2024khronosunifiedapproachspatiotemporal} jointly performs SLAM and spatio-temporal semantic mapping, separating short-term dynamics from long-term changes with a factor-graph formulation. Panoptic Multi-TSDFs~\cite{schmid2022panoptic} provide a volumetric representation for multi-resolution dynamic mapping. ObVi-SLAM~\cite{ballester2024obvi} combines visual features with an object-level map for long-term multi-session localization. POV-SLAM~\cite{qian2023povslamprobabilisticobjectawarevariational} tracks semi-static object changes using a probabilistic variational framework. 3DGS-CD~\cite{10852207} applies 3D Gaussian splatting for change detection from unaligned image sets, and Gaussian Mapping for Evolving Scenes~\cite{yugay2025gaussian} keeps a 3D Gaussian map consistent as a scene changes over time. Unlike these methods, which generally need manually collected data or are built for offline reconstruction, our approach makes \emph{autonomous, repeated} semantic updates inside a lightweight 2D-map-based framework.

\subsection{Active Perception in Dynamic Environments}
Persistent monitoring in changing environments also raises the question of when and where to revisit. Bonnevie et al.~\cite{bonnevie2021longterm} use an information gain from previous visits to revisit highly dynamic parts of the environment more often. Yu et al.~\cite{yu2023correlated} pose persistent monitoring as a correlated orienteering problem that maximizes the reward collected on a visit under a travel budget. In this work we give equal weight to all areas to follow a fixed traversal plan in each run, and focus instead on how to update the semantic map efficiently once a revisit occurs. Weighting revisits by expected change is a natural extension that we leave to future work.
\begin{figure*}
\vspace{1mm}
    \centering
\includegraphics[trim={0cm 0cm 1cm 0cm},clip,width=\linewidth]{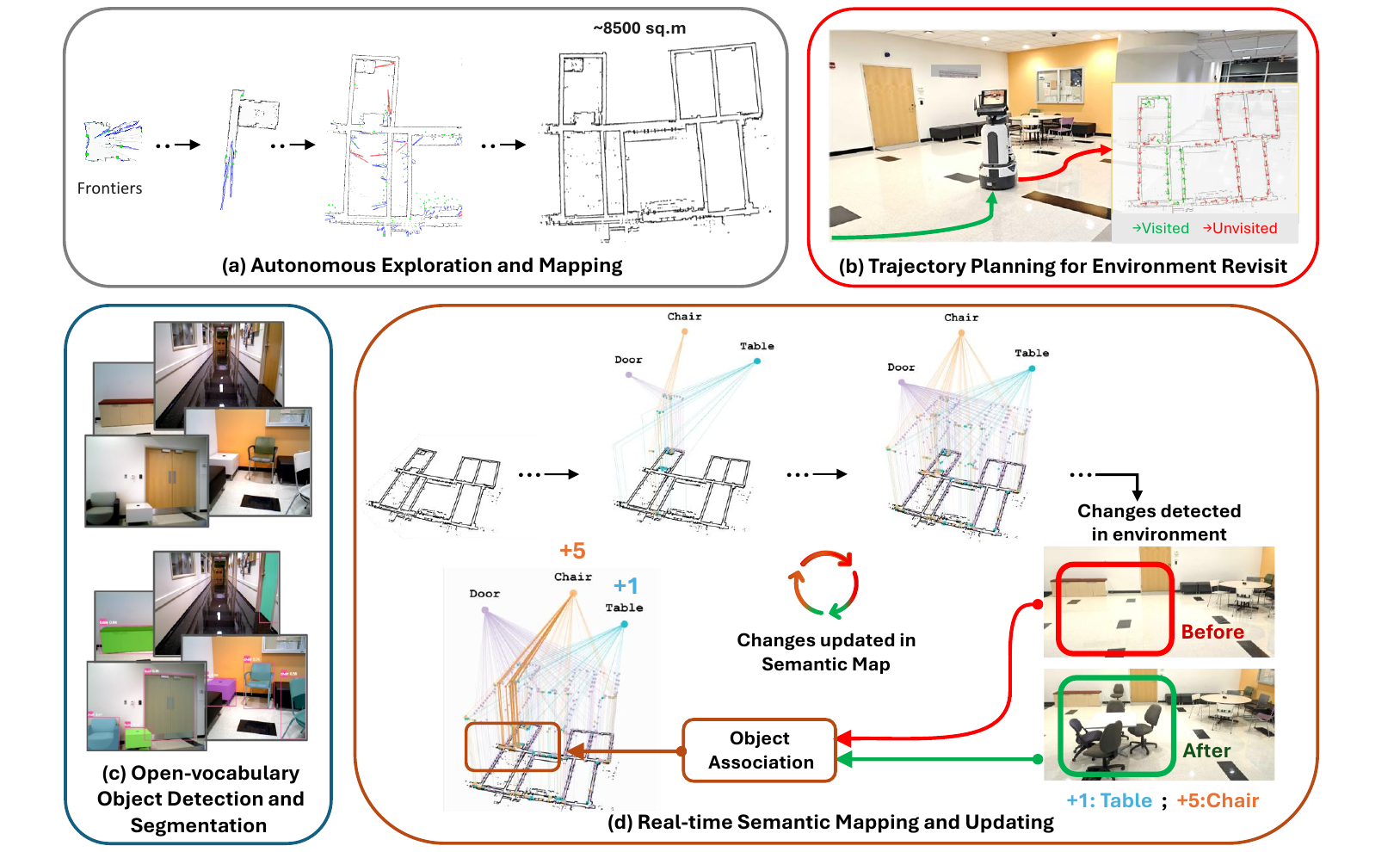}
\vspace{-1mm}
    \caption{Illustration of our system for autonomous exploration, semantic mapping and map update.}
    \label{fig:overview}
\end{figure*}

\section{System}

Our system consists of two stages. In the first stage, a mobile robot autonomously explores the environment and incrementally builds a 2D occupancy grid map. In the second stage, the robot relocalizes within the built map and performs semantic mapping by detecting, segmenting, and associating objects into a semantic object graph. An overview is presented in Fig.~\ref{fig:overview}.

\textbf{Design Rationale.} We separate geometric mapping from semantic updating for a few reasons. \emph{First}, in persistent monitoring the geometry stays largely the same and only the objects change. So running full SLAM from start on every visit repeats work that does not need to be redone. We build the geometric map once and run lightweight, relocalization-based semantic passes on each later visit, so the cost of a revisit does not grow with the number of revisits. \emph{Second}, this improves localization during semantic mapping: LiDAR SLAM such as GMapping~\cite{gmapping} gives noisy poses while the map is still being built, whereas relocalizing inside a finished map with AMCL~\cite{navigation_stack} gives more stable poses of the robot. \emph{Third}, the modular design lets us upgrade one component, such as the exploration planner or the object detector, without touching the others.
Running the semantic stage online and onboard, instead of recording each pass for offline processing, lets the robot update the map during a pass and avoids storing and replaying large image and depth streams from every visit.

\subsection{Autonomous Exploration and Map Building}
\label{exp_and_map}
Our exploration method builds on the frontier exploration module of~\cite{yamauchi1997frontier} and adds a dynamic search window for large environments. In the original method, as the map grows, far-away frontiers get a high cost and are rarely chosen, so parts of the map are left unexplored, and tuning the cost parameters for large spaces is not intuitive.


\begin{algorithm}[t]
\caption{Dynamic Window Frontier Exploration}\label{alg:dwfe}
\begin{algorithmic}[1]
\State \textbf{Input}: Occupancy map $M$, local, global radii $r_l$, $r_g$, local, global threshold $T_l$,  $T_g$, max time $T_{max}$, weights $\alpha, \beta$
\State Initialize robot position $p$, time $t \gets 0$
\While{$t < T_{max}$ }
    \State $F \gets$ frontiers within $r_l$ around $p$
    \If{$|F| < T_l$}
        \State $F \gets$ frontiers within $r_g$ around $p$
        \If {$|F| < T_g$}
            \State \textbf{break}  \Comment{Terminate if not enough frontiers}
        \EndIf
    \EndIf
    \State $f^* \gets \arg\min_{f \in F} \left( \alpha \cdot distance(p, f) + \beta \cdot \text{size}(f) \right)$
        \State \textbf{yield} $f^* $ \Comment{Output the best frontier for this iteration}
        \State Navigate to $f^*$
    \State $t \gets t + 1$
\EndWhile
\end{algorithmic}
\end{algorithm}

\textbf{Dynamic Window Frontier Exploration (DWFE).} We adjust the robot's search area between a local and a global region, as detailed in Algorithm~\ref{alg:dwfe}\footnote{Code available at \url{https://github.com/IRVLUTD/dynamic-window-frontier-exploration}}. At each step the robot looks for frontiers in the current search window and navigates to the best one. If the number of frontiers within the local radius drops below a threshold, the search expands to a large global radius. This keeps the robot moving forward and covering local regions well, which reduces redundant coverage. Conventional method~\cite{yamauchi1997frontier} does not adapt to how many frontiers are nearby, so the robot can be pulled back and forth several times. The dynamic window instead stays local while nearby frontiers remain and only widens when the local area is mostly covered. Exploration stops when the global frontier count falls below a threshold or a time limit is reached, and the map is saved. The module runs with GMapping~\cite{gmapping} (Fig.~\ref{fig:overview}(a)). We record the exploration trajectory for traversal planning later, and the saved occupancy map is used to localize the robot during semantic mapping.



\subsection{Autonomous Semantic Mapping}

We utilize the saved occupancy map and the recorded robot exploration path to plan a trajectory to traverse the environment again swiftly, to construct the semantic map.

\subsubsection{Environment Traversing}
\label{subsubsec:env_traverse}

\begin{figure}[!h]
    \centering
\includegraphics[width=\linewidth]{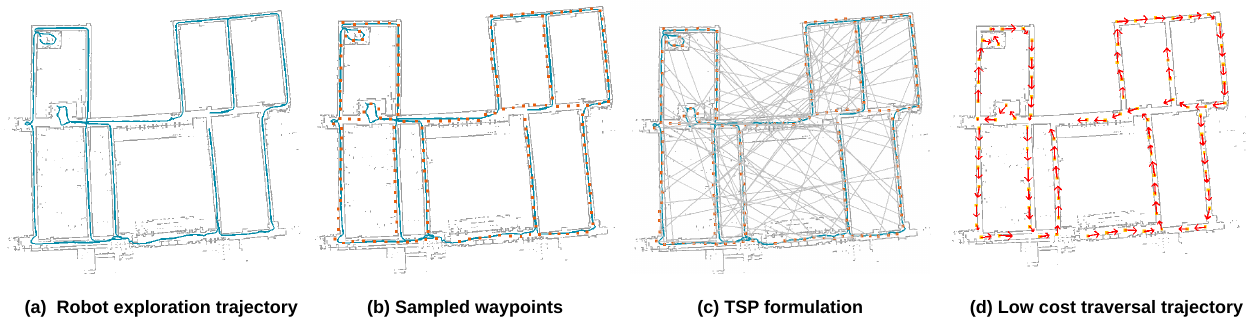}
\vspace{-3mm}
    \caption{Environment traversal trajectory planning.}
    \label{fig:traverse_plan}
\end{figure}

To build or update the semantic map, the robot revisits locations and identifies the objects there. We sample $n$ waypoints \( P = \{p_1, \dots, p_n\} \) along the recorded exploration path with at least $2$\,m spacing (Fig.~\ref{fig:traverse_plan}(a-b)), form a complete graph with Euclidean edge weights $w_{ij} = \|p_i - p_j\|$ (Fig.~\ref{fig:traverse_plan}(c)), and order them with a greedy nearest-neighbor heuristic for the Traveling Salesman Problem: starting from $p_1$, we repeatedly move to the closest unvisited waypoint (Fig.~\ref{fig:traverse_plan}(d)). This is not optimal, but it gives efficient trajectories in practice and runs in $O(n^2)$ time.

\subsubsection{Object Detection and Segmentation}
\label{subsubsec:obj_det}


During traversal, object detection and segmentation run in real time to build the semantic map. This stage only assumes two interchangeable components: an open-vocabulary object detector that returns a category label and a bounding box for each object in the robot's RGB image, and a promptable segmentation model that returns the object mask within this bounding box. Any detector and segmenter that meet this interface can be used, so the system is not tied to a specific model. The detector should be open-vocabulary so that new object categories can be added at deployment time without retraining. The specific models we use in our experiments are given in Sec.~\ref{sec:experiments}.


For each segment, we back-project the masked depth pixels with the camera intrinsics to get a 3D point cloud \( \mathbf{P}_{\text{camera}} \) in the camera frame. We transform it to the map frame $M$ using the camera pose in the robot base frame \( \mathbf{T}_{\text{camera}} \in \mathbb{SE}(3) \) and the robot pose in the map frame \( \mathbf{T}_{\text{robot}} \in \mathbb{SE}(3) \): \( \mathbf{P}_{\text{map}} = \mathbf{T}_{\text{robot}} \cdot \mathbf{T}_{\text{camera}} \cdot \mathbf{P}_{\text{camera}} \). The mean of this point cloud is the object's 3D position in the map. Although navigation uses a 2D occupancy grid, object positions are computed and stored in full 3D, so objects at different heights get distinct centroids and stay separable in the object graph.



\subsubsection{Semantics Representation and Mapping}

Our semantic map is a hybrid representation. At the lower level, we use a 2D occupancy grid map $M$ that captures the free space and obstacles in the environment. At the higher level, we build a \emph{semantic object graph} defined by a set of nodes \( V = \{v_1, v_2, \dots, v_n\} \), where each node \( v_i \) is one uniquely detected object (Fig.~\ref{fig:overview}(d)). We use the term object graph rather than topological map, since the nodes represent objects and not the connectivity of navigable space. Each node $v_{i}$ has the following attributes.
\begin{itemize}
\item $v_{i}.\text{id}$: A unique id of the object
    \item $v_{i}.\text{category}$: Object category label
    \item $v_{i}.\text{position}$: Mean 3D position of the object computed in the reference frame of the map via full SE(3) transforms. This attribute links the object graph to the occupancy map in a hierarchical way.
    \item $v_{i}.\text{confidence}$: the confidence score of the detected object from object detection
\end{itemize}
This representation facilitates fast object association and easy update of nodes during environment changes.


\begin{algorithm}[!h]
\caption{Semantic Mapping and Updating}\label{alg:sem_map_update}
\begin{algorithmic}[1]
\State \textbf{Input}: Traversal trajectory $T_s$, Distance thresholds $\Delta = \{\delta_c \mid c \in \text{categories}\}$
\State \textbf{Output}: Updated semantic map $G_T(V')$
\If{no prior map exists}
    \State Initialize $V' \gets \emptyset$, id $\gets 0$
\Else
    \State Load $G_T(V)$
    \State $V' \gets V$, \quad $\text{id} \gets \max_{v \in V'} v.\text{id} + 1$
\EndIf
\While{Traversing $T_s$}
\State $\mathcal{O}_{\text{current}} \gets$ objects detected in FoV
\State $\mathcal{V}_{\text{expected}} \gets$ expected nodes in FoV from $G_T(V')$
\State $\mathcal{O}_{\text{unmatched}} \gets \mathcal{O}_{\text{current}}$ \Comment{Initialize unmatched objects}
\For{$v \in \mathcal{V}_{\text{expected}}$}
            \If{no $o \in \mathcal{O}_{\text{current}}$ such that \\
            \hspace{1em} $o.\text{category} = v.\text{category}$ and \\
            \hspace{1em} $\text{dist}(P_{\text{mean}}(o), v.\text{position}) \leq \delta_{v.\text{category}}$}
                \State \colorbox{red!16}{\strut $V' \gets V' \setminus \{v\}$} \Comment{Remove node}
            \Else
            \State $\mathcal{O}_{\text{unmatched}} \gets \mathcal{O}_{\text{unmatched}} \setminus \{o\} $
            \EndIf
        \EndFor
\For{$o \in \mathcal{O}_{\text{unmatched}}$}
    \State $V_c \gets \{v \in V' \mid v.\text{category} = o.\text{category}\}$
    \State $d_{\min} \gets \infty$
        \If{$V_c \neq \emptyset$}
            \State $d_{\min} \gets \min_{v \in V_c} \text{dist}(P_{\text{mean}}(o), v.\text{position})$
        \EndIf
        \If{$d_{\min} > \delta_{o.\text{category}}$}
            \State $v_{new} =  (\text{id}, o.\text{category}, P_{\text{mean}}(o.\text{segment}),$ 
            \State \hspace{3.25cm}\qquad $o.\text{confidence})$
            \State \colorbox{green!20}{\strut $V' \gets V' \cup \{v_{new}\}$} \Comment{Add new node to $V'$}

            \State $\text{id} \gets \text{id} + 1$
        \EndIf
\EndFor
\EndWhile
\State \textbf{Return} $G_T(V')$
\end{algorithmic}
\end{algorithm}

\textbf{Semantic Mapping and Updating}. To build the semantic map, we first load the occupancy map $M$ built as detailed in Sec.~\ref{exp_and_map} and localize the robot within it. If the semantic mapping process is performed for the first time, an object graph is initialized with an empty node set $V' = \emptyset$. Otherwise, the previously constructed semantic map is loaded ( $V'  \gets V$). We match a detection to an existing node when they have the same category and the centroid distance is below a threshold $\delta$. The categories come from the open-vocabulary detector at run time and are not fixed in advance. By default we use a single distance $\delta$ for all categories. When finer separation is needed, $\delta$ can be set per category, for example near the typical object size, so that small and large objects need not share one value.

We use centroid distance for association because it stays stable across revisits. With IoU-based matching, the mask of an object changes when the object is placed in a different pose, or when the robot views it from a different direction on another run. The overlap between the new mask and the stored one can be low even for the same object. Appearance-feature matching has a similar issue, since it needs a similar viewpoint to give a reliable match. The centroid of an object moves much less under these conditions, which makes it a more reliable cue for matching the same object across visits.

\begin{figure}[!h]
\vspace{-2mm}
    \centering
    
\includegraphics[trim={0cm 0cm 0cm 0.2cm},clip,width=\linewidth]{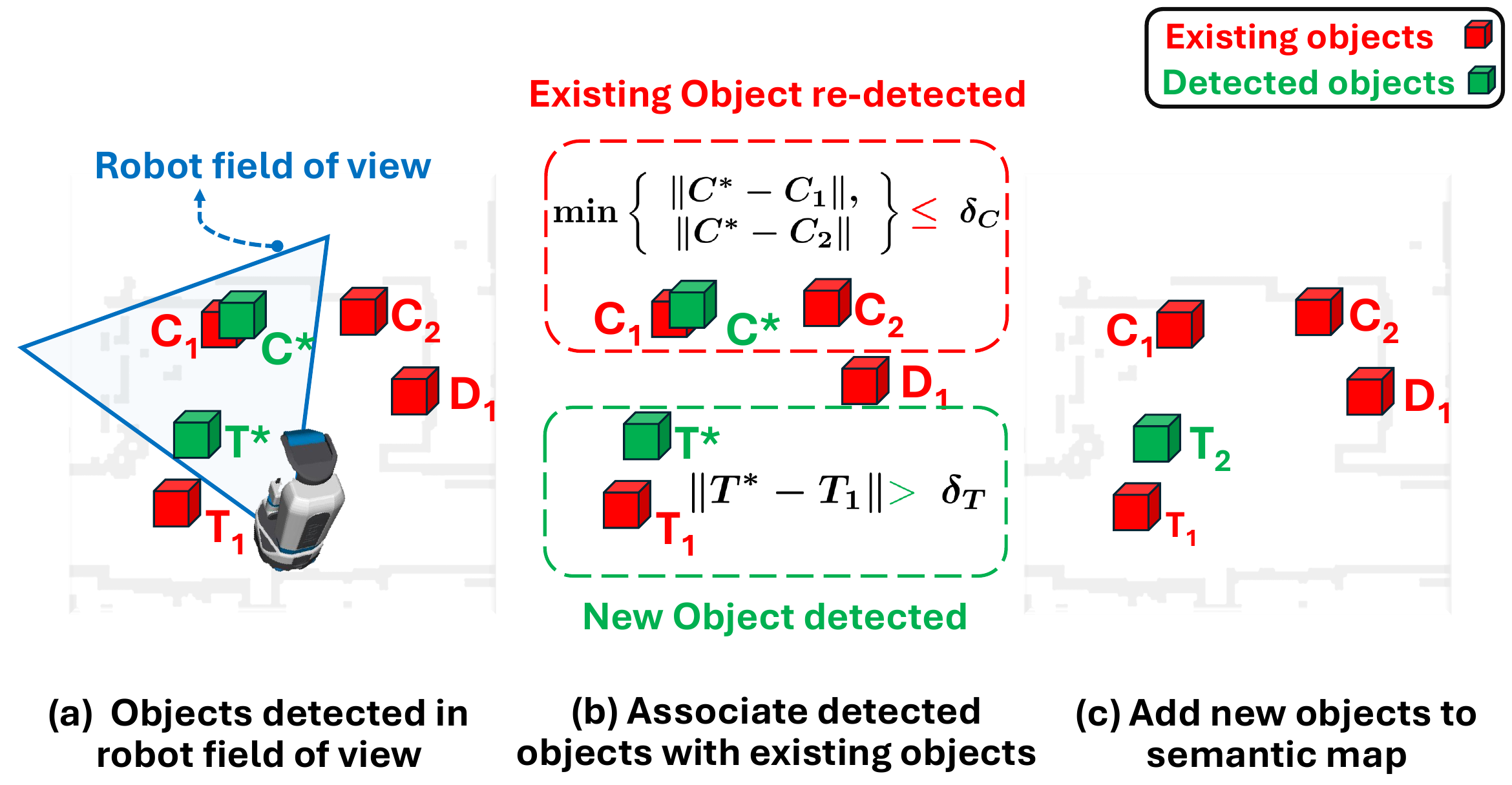}
    \vspace{-4mm}
    \caption{Updating Semantic Map through Object Association.}
    \label{fig:add_semmap}
\end{figure}

Once initialized, the robot follows the traversal trajectory $T_s$ (Sec.~\ref{subsubsec:env_traverse}). At the same time, the detector and segmenter process the current view to find objects $\mathcal{O}_{\text{current}}$ and compute their map positions (Fig.~\ref{fig:add_semmap}(a)); these are initially unmatched, $\mathcal{O}_{\text{unmatched}}$. At each step we form an approximate field-of-view (FoV) polygon from the limits of the current depth point cloud and query the map for the nodes $\mathcal{V}_{\text{expected}}$ that should be visible. Alg.~\ref{alg:sem_map_update} then updates the node set $V'$.

\vspace{0.7mm}
\textbf{Node Removal.} For each expected node $ v \in \mathcal{V}_{\text{expected}} $, the algorithm checks for a matching object $ o \in \mathcal{O}_{\text{current}} $ of the same category ($ o.\text{category} = v.\text{category} $) and within the threshold distance $ \text{dist}(P_{\text{mean}}(o), v.\text{position}) \leq \delta_{v.\text{category}} $. If no match is found, the node $ v $ is removed from $ V' $, indicating that the corresponding object is no longer present in the expected location (e.g., moved or removed). The matched objects are removed from $ \mathcal{O}_{\text{unmatched}}$. Crucially, nodes are only evaluated for removal when they fall within the robot's current field of view; objects outside the FoV are preserved by default, providing implicit robustness against transient detection failures or temporary occlusions in unobserved regions.

\textbf{Node Addition.} For each unmatched object $ o \in \mathcal{O}_{\text{unmatched}} $, the algorithm identifies nodes $ V_c = \{v \in V' \mid v.\text{category} = o.\text{category}\} $ belonging to the same category. If no such category exists $ V_c = \emptyset $ or an object is not associated with existing ones $ \min_{v \in V_c} \text{dist}(P_{\text{mean}}(o), v.\text{position}) > \delta_{o.\text{category}} $, a new node $ v_{new} = (\text{id}, o.\text{category}, P_{\text{mean}}(o.\text{segment}), o.\text{confidence}) $ is added to $ V' $.
The object association mechanism is illustrated in Fig.~\ref{fig:add_semmap}(b).
This process continues until the robot finishes the traversal of $T_s$. By updating the node set $V'$ as it moves, the algorithm captures both intra-session changes, which happen during a single traversal after the robot has already passed a section, and inter-session changes, which build up across separate revisits. The resulting map combines the geometry of the occupancy map with the object-level information in the object graph, which supports fast scene understanding.


\section{Experiments}
\label{sec:experiments}
We instantiate the detector and segmenter from Sec.~\ref{subsubsec:obj_det} with GroundingDINO~\cite{groundingdino} as the open-vocabulary detector and MobileSAM~\cite{mobilesam}, a faster version of SAM~\cite{kirillov2023sam}, as the promptable segmenter. Any comparable models could be used in their place.

\subsection{Experimental Setup}
\textbf{Environments.} We evaluate semantic mapping in two real-world environments with different object categories.

\textbf{Environment A} is a large indoor space of about $\mathbf{93m \times 90m}$, with 18 corridors and a total traversal length of about \textbf{800\,m}. The corridors are furnished with typical office furniture, so we map three classes: \textbf{table}, \textbf{door}, and \textbf{chair}.

\textbf{Environment B} is a smaller laboratory of $9m \times 13m$ with a more diverse set of objects: \textbf{televisions, trash bins, umbrellas, chairs, cabinets}, and \textbf{persons}. Only objects with detection confidence $\geq 0.75$ are considered.

\textbf{Robot and Hardware.} We use a Fetch mobile robot with a LiDAR sensor and an RGB-D camera, equipped with a laptop with an RTX 4090 GPU running Ubuntu 20.04 and ROS Noetic.

\subsection{Autonomous Exploration and Map Building}

\textbf{Simulation Benchmark}. To evaluate the Dynamic Window Frontier Exploration method (Alg.~\ref{alg:dwfe}), we ran experiments in a large simulated hospital environment~\cite{aws_robomaker_hospital_world} of about $1440~\text{sq.m}$ in Gazebo. We ran 20 exploration trials, each spawning a Fetch robot at a different start location for 15 minutes. The same start locations are used for our method and the conventional frontier baseline~\cite{Horner2016}, and we record the area covered and the path length traveled for a direct comparison.

\begin{table}[!h]
\vspace{1mm}
\centering

\renewcommand{\arraystretch}{2} 

\resizebox{\linewidth}{!}{
\begin{tabular}{|c|c|c|}
\hline 
\textbf{Method} &
\textbf{Average Area Covered (sq.m)} $\boldsymbol{\uparrow}$ &
\textbf{Average Path Length Traveled (m)} $\boldsymbol{\downarrow}$ \\
\hline

Baseline~\cite{Horner2016}
& 669.12
& 107.10 \\
\hline

\textbf{DWFE (ours)}
& 721.05
& 94.30 \\
\hline

\end{tabular}
}
\vspace{-1mm}
\caption{Average area coverage and path length for baseline vs. DWFE in a large-scale simulated environment.}
\label{tab:exploration_results}
\vspace{-3mm}
\end{table}

 The results in Table~\ref{tab:exploration_results} show that our method is more efficient than this baseline. Compared to the baseline, it explores $\mathbf{\approx 7.8\%}$ more area while traveling $\mathbf{\approx 12\%}$ less on average. We report this as an internal validation on the exploration module rather than a benchmark against state-of-the-art planners. Since the module is interchangeable, a stronger planner can be utilized without changing the rest of the system. 

 \textbf{Real-World Execution}. We integrate mapping, exploration, and navigation to map environments autonomously. GMapping~\cite{gmapping} builds a 0.1\,m/pixel occupancy map from laser scans, while DWFE selects frontiers and \textit{move\_base}~\cite{navigation_stack} drives the robot to them with global and local planning and obstacle avoidance. The robot is capped at 0.6\,m/s with a 0.7\,m inflation radius. In Environment A, 150\,min of exploration (including stops for people in narrow corridors) produced a full map and 2{,}250 trajectory points at 0.25\,Hz. In Environment B, exploration finished in about 4\,min with 212 points at 1\,Hz.
 
\textbf{Environment Traversing.} Using the greedy TSP from Sec.~\ref{subsubsec:env_traverse}, traversal trajectories were generated from the recorded exploration poses: 130 poses for Environment A and 15 poses for Environment B. Revisiting was significantly faster, taking approximately 35\,min for Environment A and 2\,min for Environment B, demonstrating the efficiency of the optimized traversal.

\subsection{Semantic Mapping and Updating}

\begin{figure}[ht]
    \centering
    \includegraphics[width=\linewidth]{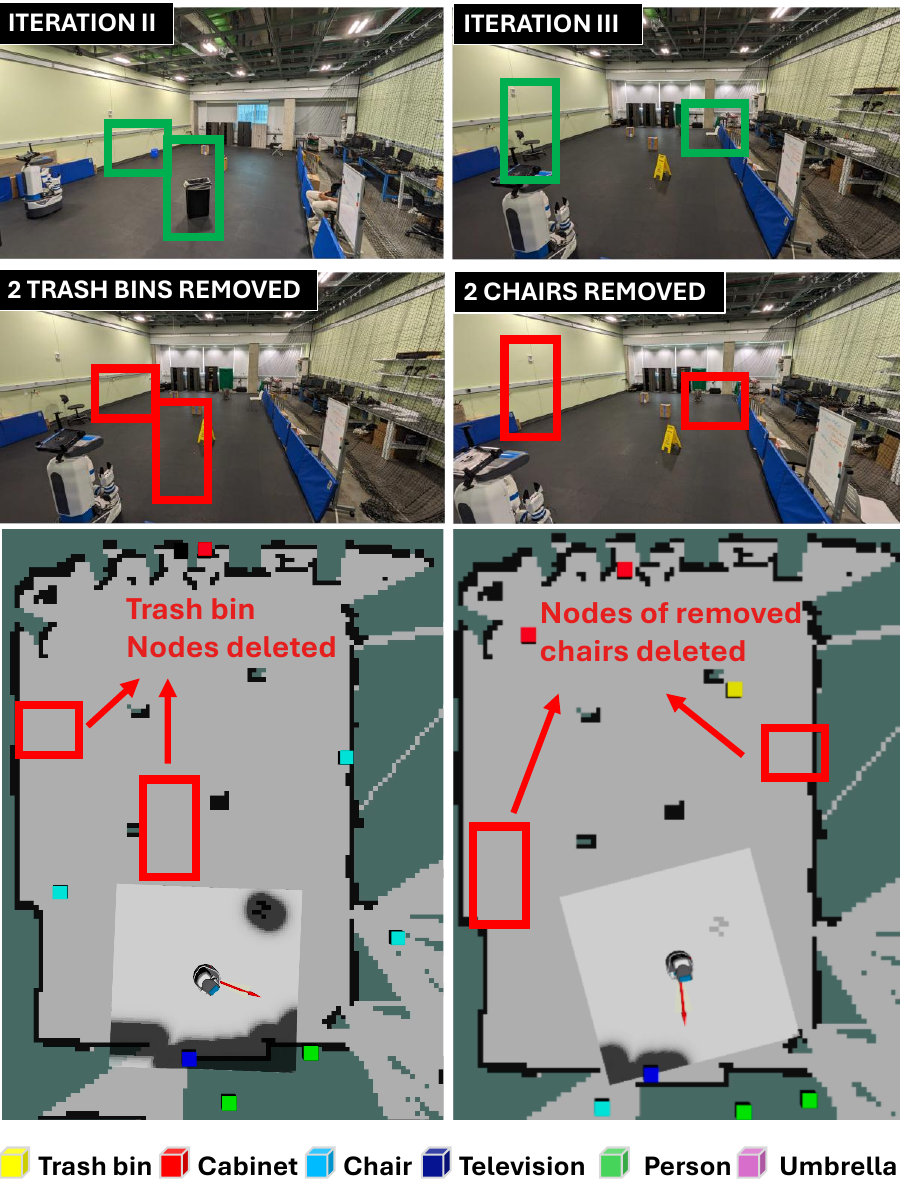}
    \vspace{-3mm}
    \caption{Visualization of environmental changes in Environment B: Iteration II (2 trash bins removed) and Iteration III (2 chairs removed). When robot revisited, these changes are reflected in the updated semantic map.}
    \label{fig:sem-map-update}
\end{figure}

\textbf{Semantic Map Construction.} The system loads the built occupancy map and localizes using AMCL~\cite{navigation_stack}, then progressively builds the semantic map. We set the object association distance threshold to 0.7\,m for all categories. Two additional iterations are then performed, during which the environment is modified through removal, addition, and relocation of objects. Fig.~\ref{fig:sem-map-update} shows changes in Environment B reflected during semantic updating. 

\textbf{Object Classes.} We set a detection confidence threshold of 0.75 for GroundingDINO~\cite{groundingdino}. In Environment A, only three object classes (chairs, tables, and doors) were consistently mapped above this threshold, as these are the main static objects on the floor. We excluded \textit{human} category as they are high dynamic and produce unreliable ground truth. No other categories were regularly present in the environment. This reflects the make-up of the environment and not a limit of the method, since the open-vocabulary detector can recognize new categories without retraining. Environment B has a more diverse set of object classes such as cabinets, televisions, and trash bins mapped using a similar threshold.

\vspace{1mm}

\textbf{Evaluation Protocol.} We report results at the object level. For each real-world iteration we record the ground-truth objects that are physically present in the environment, counted by hand. An object in the map is counted as a true positive if there is a ground-truth object of the same category whose real position is within the association distance of the mapped node. A mapped node with no matching ground-truth object is a false positive, and a ground-truth object with no matching node is a false negative. From these counts we report precision, recall, and F1 per class. We count objects, rather than reconstruction quality, because the goal of the system is to keep an accurate object-level map across visits, which is why we also report false positives and false negatives.

\begin{table}[!h]
\centering

\renewcommand{\arraystretch}{1.2}      
\setlength{\extrarowheight}{2pt}       
\setlength{\tabcolsep}{5pt}

\resizebox{\linewidth}{!}{
\begin{tabular}{|l|l||c|c||c|c||c|c|}
\hline
\textbf{Env} & \textbf{Class} 
& \multicolumn{2}{c||}{\textbf{Iteration I (Initial run)}} 
& \multicolumn{2}{c||}{\textbf{Iteration II (vs I)}} 
& \multicolumn{2}{c|}{\textbf{Iteration III (vs I)}} \\
\cline{3-8}
& & \textbf{\hspace{3mm} GT \hspace{3mm}} & \textbf{Map} 
& $\Delta$\textbf{GT} & $\Delta$\textbf{Map} 
& $\Delta$\textbf{GT} & $\Delta$\textbf{Map} \\
\hline

\multirow{3}{*}{\textbf{A}}  
& Table      & 45  & 45  & 0   & -6  & 0* & -8  \\ \cline{2-8}  
& Chair      & 109 & 84  & -10 & -11 & 0* & -15 \\ \cline{2-8}  
& Door       & 170 & 153 & 0   & +2  & 0  & +3  \\  
\hline

\multirow{6}{*}{\textbf{B}}  
& Person     & 2 & 2 & 0* & 0 & 0 & 0 \\ \cline{2-8}
& Trash Bin  & 2 & 2 & -2 & -2 & -2 & -1 \\ \cline{2-8}
& Umbrella   & 1 & 1 & -1 & -1 & -1 & -1 \\ \cline{2-8}
& Chair      & 3 & 3 & 0 & 0 & -2 & -2 \\ \cline{2-8}
& Television & 1 & 2 & 0 & -1 & 0 & -1 \\ \cline{2-8}
& Cabinet    & 3 & 2 & 0 & -1 & 0 & 0 \\  
\hline

\end{tabular}
}

\caption{Semantic map object (obj) counts with Iteration I as reference.
\textbf{GT}: Objs present in the environment during Iteration I.
\textbf{Map}: Objs mapped in Iteration I.
$\Delta$\textbf{GT}: Change in real-world objs relative to Iteration I.
$\Delta$\textbf{Map}: Change in mapped objs relative to Iteration I.
* indicates relocation.}
\label{tab:result-num-obj}
\end{table}

\textbf{Quantitative Analysis.} 
Table~\ref{tab:result-num-obj} reports the semantic mapping results across three iterations, with Iteration~I serving as the reference. For Iteration~I, we show the absolute ground truth (GT) objects and those mapped by the robot (Map). For Iterations~II and III, we report only the relative changes ($\Delta$GT, $\Delta$Map) compared to Iteration~I.  

In \textbf{Environment A}, tables and doors stayed stable across iterations, while chairs were harder because of occlusions from people. In Iteration~II, $\Delta$GT = --10 reflects ten chairs removed after Iteration~I, and the mapped count drops by a matching $\Delta$Map = --11. In Iteration~III, the chairs had $\Delta$GT = 0* (ten chairs relocated) but a larger mapped change ($\Delta$Map = --15), and tables had $\Delta$GT = 0 but $\Delta$Map = --6. These mismatches happened because some relocated chairs occluded others, and some were placed along the robot's path $T_s$, so the robot took a detour to avoid them and missed nearby chairs and tables. This raised the false negatives, as shown in Table~\ref{tab:result-det-combined}.

In \textbf{Environment B}, detections were stable, with most categories at $\Delta$GT = 0 and $\Delta$Map = 0. Small deviations occurred for cabinets and televisions, where dark color or low lighting caused occasional missed detections.
Overall, stable objects such as tables and doors stay consistent across iterations, while occlusion-prone categories such as chairs vary more. The F1 scores (0.75-0.87 for chairs, 0.80-0.87 for doors and tables) are in the range reported for open-vocabulary detectors in cluttered indoor scenes~\cite{groundingdino}. The drop in chair F1 from 0.87 to 0.75 comes mainly from human occlusions and relocated objects blocking the planned path, not from a failure of the detection pipeline.
These results demonstrate that the proposed system maintains reliable semantic mapping across multiple iterations despite dynamic changes. Future work will focus on improving resilience under occlusion and low-lighting conditions.





\begin{table}[!h]
\centering
\resizebox{\linewidth}{!}{
\renewcommand{\arraystretch}{1.3}
\setlength{\tabcolsep}{5pt}
\begin{tabular}{|l||c|c|c|c||c|c|c|c||c|c|c|c|}
\hline
\textbf{Class} & \multicolumn{4}{c||}{\textbf{Iteration I}} & \multicolumn{4}{c||}{\textbf{Iteration II}} & \multicolumn{4}{c|}{\textbf{Iteration III}} \\
\cline{2-5} \cline{6-9} \cline{10-13}
               & \textbf{TP} & \textbf{FP} & \textbf{FN} & \textbf{F1} & \textbf{TP} & \textbf{FP} & \textbf{FN} & \textbf{F1} & \textbf{TP} & \textbf{FP} & \textbf{FN} & \textbf{F1} \\
\hline
\multicolumn{13}{|c|}{\textbf{Environment A}} \\
\hline

Table         & 36  & 9  & 9  & 0.80  & 36  & 3  & 9  & 0.86  & 33  & 4  & 12  & 0.80  \\ \hline
Chair         & 84  & 0  & 25  & 0.87  & 72  & 1  & 27  & 0.84  & 67  & 2  & \textbf{42 } & 0.75  \\  \hline
Door          & 140 & 13 & 30  & 0.87  & 140 & 15 & 30  & 0.86  & 142 & 14 & 28  & 0.87  \\  
\hline
\multicolumn{13}{|c|}{\textbf{Environment B}} \\
\hline
Person        & 2   & 0   & 0   & 1.00  & 2   & 0   & 0   & 1.00  & 2   & 0   & 0   & 1.00\\  \hline
Trash Bin     & 2   & 0   & 0   & 1.00  & 0   & 0   & 0   & -   & 0   & 1   & 0   & 0.00 \\  \hline
Umbrella      & 1   & 0   & 0   & 1.00  & 0   & 0   & 0   & -   & 0   & 0   & 0   & - \\  \hline
Chair         & 3   & 0   & 0   & 1.00  & 3   & 0   & 0   & 1.00  & 1   & 0   & 0   & 1.00\\ \hline
Television    & 1   & 1   & 0   & 0.67 & 1   & 0   & 0   & 1.00  & 1   & 0   & 0   & 1.00\\  \hline
Cabinet       & 2   & 0   & 1   & 0.80 & 1   & 0   & 2   & 0.50 & 2   & 0   & 1   & 0.80 \\  
\hline

\end{tabular}
}
\caption{Semantic map object count across Iteration I, II \& III for A \& B Environments. \textbf{TP}: True Positives, \textbf{FP}: False Positives, \textbf{FN}: False Negatives, \textbf{F1} scores are reported.}
\vspace{-2mm}
\label{tab:result-det-combined}
\vspace{-5mm}
\end{table}

\begin{figure}[!h]
    \centering
    \includegraphics[width=0.9\linewidth]{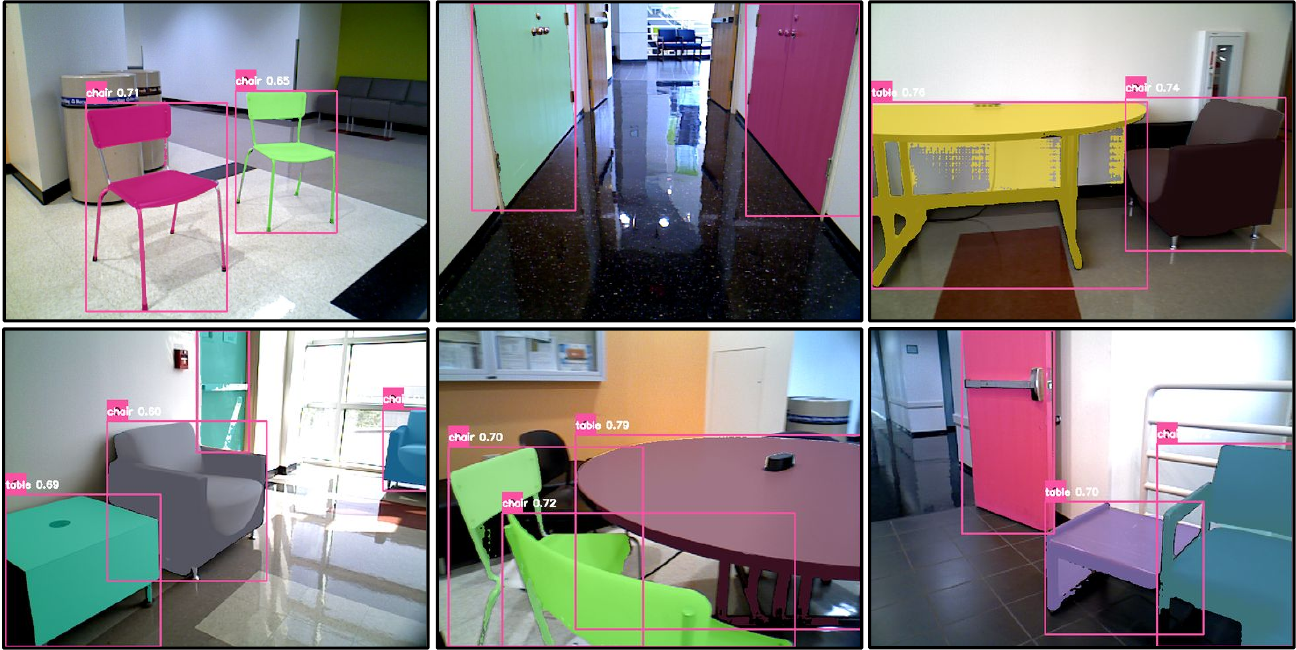}
    \caption{Examples of accurate detection and segmentation of chairs, tables and doors in Environment A.}
    \label{fig:detection}
    \vspace{-3mm}
\end{figure}

\textbf{Qualitative Analysis.} Fig.~\ref{fig:detection} illustrates several examples of accurate object detection during our experiments.

\label{subsec:qual_analysis}

\begin{figure}[H]
\vspace{0.2mm}
    \centering
\includegraphics[width=0.9\linewidth]{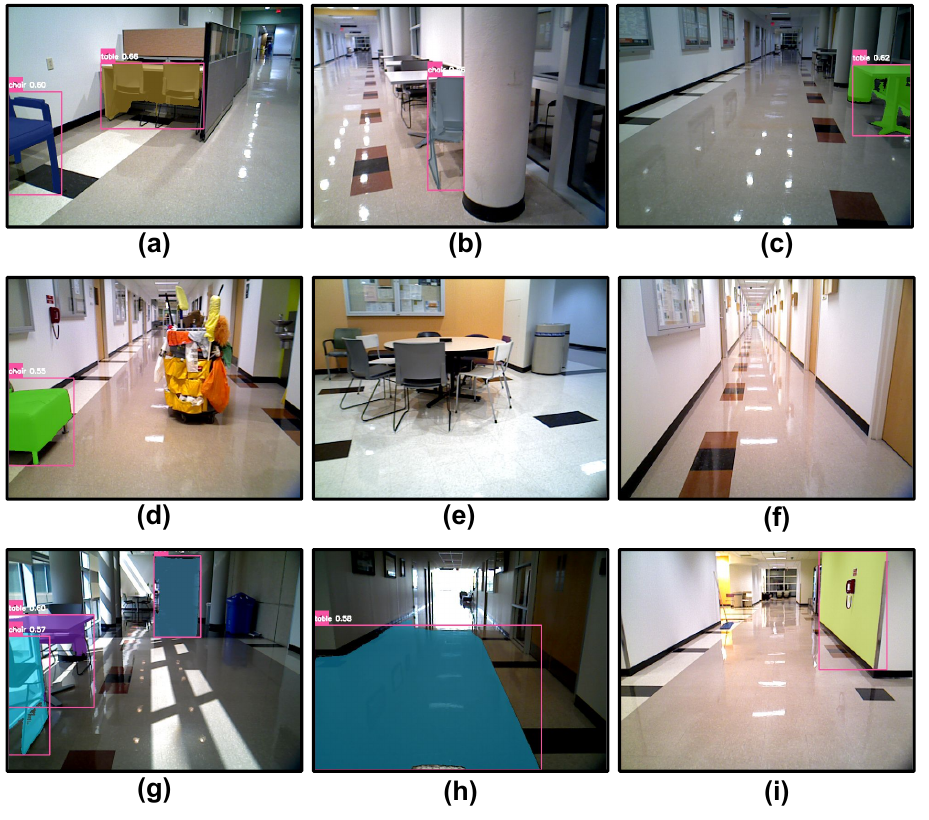}
    \vspace{-2mm}
    \caption{Representation of various challenging scenarios involved in object detection and semantic mapping.}
    \label{fig:challenges}
    \vspace{-2mm}
\end{figure}

In Fig.~\ref{fig:challenges}, we show several challenging scenarios in object detection and segmentation, while robot is navigating.

\textbf{Occlusion Challenges}. Close proximity of objects such as tables and chairs causes occlusion and leads to missed detections or incorrect segmentation, as seen in Fig.~\ref{fig:challenges}(a). In Fig.~\ref{fig:challenges}(c) table, chair, and a portion of the pillar are segmented together as table. The scenario seen in Fig.~\ref{fig:challenges}(b), where a chair is partially occluded by the pillar, leads to incorrect object associations when the chair is detected again at another view. 

\textbf{Detection Threshold}. Lowering the detection threshold increases false positives, while a higher threshold may miss objects. In Fig.~\ref{fig:challenges}(e), chairs in the foreground are undetected at a 0.8 threshold. careful choice of threshold for object class is important for accurate semantic mapping.


\textbf{Limitations in Constrained Spaces.}
Narrow corridors limit the robot's view of objects like doors, as seen in Fig.~\ref{fig:challenges}(f). Several of such doors are not detected, leading to a reduction in the number of object nodes in the semantic map. This is evident in Table~\ref{tab:result-num-obj}.



\textbf{Deviation from the Traversal Trajectory.}
In Fig.~\ref{fig:challenges}(d), a large bin in the corridor occluded objects, forcing the robot to take a detour. This detour resulted in the robot not seeing some objects. Such scenarios impact the quality of the semantic mapping, reducing the ability to accommodate environmental changes. 

\textbf{Impact of Lighting.}
Varying lighting conditions can cause shadows and reflections. As seen in Fig.~\ref{fig:challenges}(g) and Fig.~\ref{fig:challenges}(h), these shadows on the floor were mistook for a door and a table, respectively, reducing the overall detection and semantic map accuracy. In Fig.~\ref{fig:challenges}(c), reduced brightness at night degraded detection and segmentation.

\textbf{Structural Misidentification.}
Walls resembling doors due to shape or added features (e.g., support bars) often cause misidentification, as seen in Fig.~\ref{fig:challenges}(i). This scenario also occurred in Environment B, update II phase, where a cardboard box was identified as a trash bin, leading to a false positive.

\subsection{Comparison with Baseline}
\begin{figure}[ht]
    \centering
    \vspace{-2mm}
    \includegraphics[trim={0cm 1cm 1cm 1cm},clip,width=0.7\linewidth]{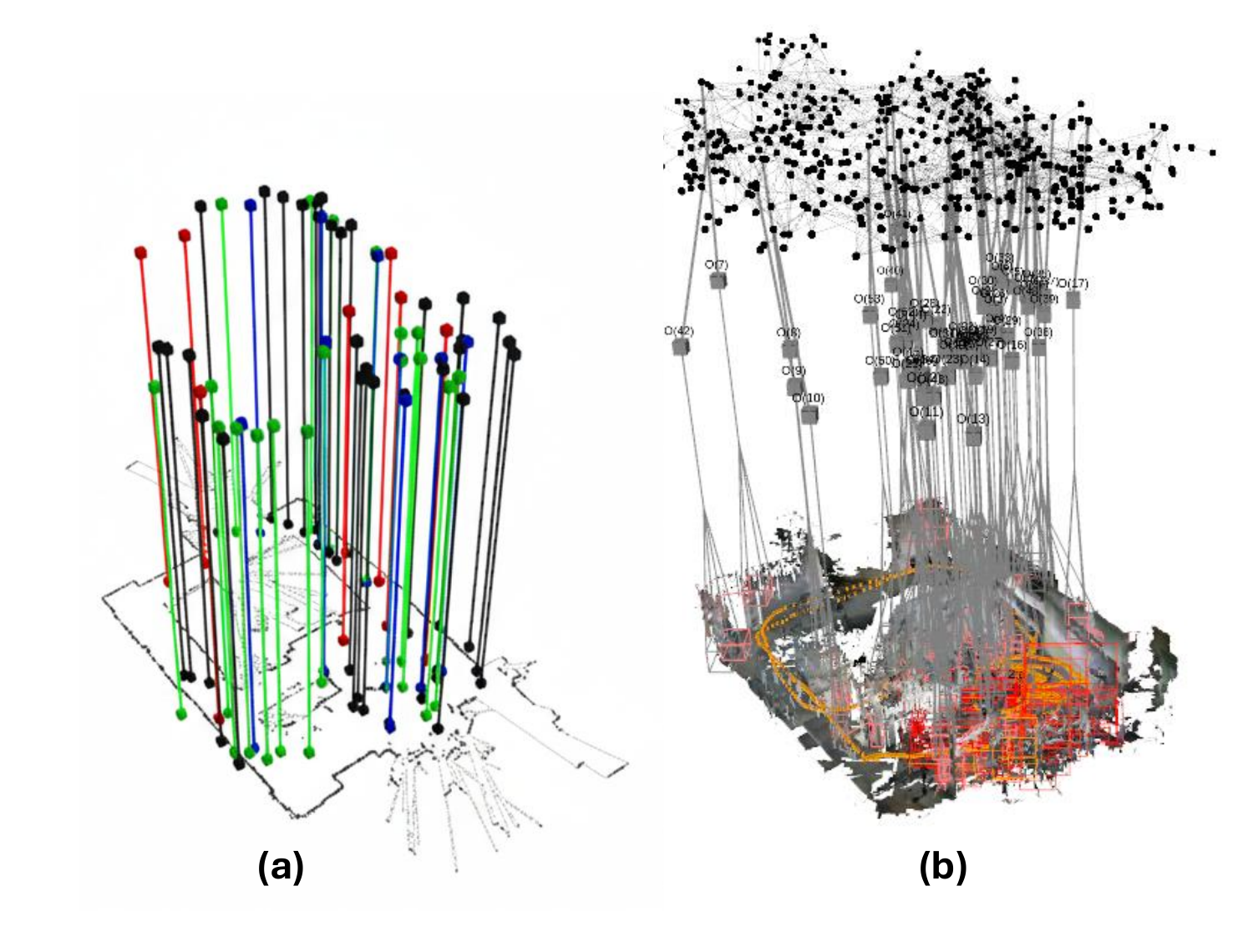}
    \vspace{-2mm}
    \caption{Intermediate semantic maps generated by : (a) our method, (b) Khronos \cite{schmid2024khronosunifiedapproachspatiotemporal}, running at 21FPS of data replay.}
    \label{fig:map_compare}
    \vspace{-2mm}
\end{figure}

We compare the object-level mapping throughput of our method with Khronos~\cite{schmid2024khronosunifiedapproachspatiotemporal} on their real-world mezzanine scene dataset, evaluated on a workstation with an NVIDIA RTX A5000 GPU. Khronos performs joint spatio-temporal metric-semantic SLAM and builds a much richer dense 3D mesh. Thus, this comparison is about how many objects each system maps in real time at a given input rate, not about reconstruction quality, where Khronos is stronger. Fig.~\ref{fig:map_compare} shows the maps from both methods at an intermediate stage.

 Table~\ref{tab:compare_with_khronos} presents the number of objects mapped. At 21 FPS replay, Khronos maps 87 objects and our method maps a comparable 86. At 30 FPS, our method maps 79 while Khronos maps 3. Khronos uses an anytime frontend that drops frames when it cannot keep up. This doesn't mean it fails to detect objects. Instead, its dense reconstruction and loop-closure optimization leave little room to register objects before frames are dropped, while our lighter pipeline keeps up. We report this as a throughput gap under a fixed compute budget, not as a claim that our mapping is more complete.


\begin{table}[!h]
\centering
\resizebox{0.7\linewidth}{!}{
\renewcommand{\arraystretch}{1.3} 
\begin{tabular}{|c|c|c|}
\hline
\multirow{2}{*}{\textbf{Data Replay Rate}} & \multicolumn{2}{c|}{\textbf{Objects Mapped by Method}} \\ \cline{2-3} 
 & \textbf{Khronos \cite{schmid2024khronosunifiedapproachspatiotemporal}} & \textbf{Ours} \\ \hline
21 FPS & 87 & 86 \\ \hline
30 FPS & 3  & 79 \\ \hline
\end{tabular}
}
\caption{Number of objects mapped by each method at different data FPS on the real-world mezzanine dataset.}
\label{tab:compare_with_khronos}
\vspace{-5mm}
\end{table}

\section{Conclusion and Future Work}
\vspace{-1mm}

We presented a modular two-stage system for persistent semantic monitoring of indoor environments. By separating geometric exploration from semantic updating, the system builds a geometric map once and then runs efficient, relocalization-based semantic passes on later visits. The semantic object graph, maintained through category and distance based association, captures both intra-session changes within a traversal and inter-session changes across revisits. Real-world experiments on a Fetch robot in environments up to $8{,}500$\,m$^2$ show effective semantic mapping and updating, with F1 scores of 0.75-0.87 across multiple iterations.

The system also has clear limitations. The distance-threshold association cannot tell a missed detection apart from a real removal, so it does not separate evidence of absence (free space where an object used to be) from absence of evidence (the detector simply failing). Adding a probabilistic persistence model would make removals more robust. The two-stage design also means the robot does not use semantic goals while exploring, and it does not actively re-plan toward changing objects during a pass. Future work will look at interactive perception, where the robot repositions itself to confirm uncertain detections, and at traversal planning that revisits regions with higher expected change first.

\section*{Acknowledgment} This work was supported by the DARPA Perceptually-enabled Task Guidance (PTG) Program under contract number HR00112220005, the Sony Research Award Program, the National Science Foundation (NSF) under Grant Nos. 2346528 and 2520553, and the NVIDIA Academic Grant Program Award. The work of T. Summers was supported by the United States Air Force Office of Scientific Research under Grant FA9550-23-1-0424 and the National Science Foundation under Grant ECCS-2047040. We would like to thank our colleague, Jishnu Jaykumar P, for his assistance during the experiments.


\bibliographystyle{ieeetr}
\bibliography{references}

\end{document}